\documentclass[11pt,a4paper]{article}
\usepackage[hyperref]{acl2020}
\usepackage{times}
\usepackage{latexsym}

\usepackage{amsfonts}
\usepackage{times}
\usepackage[pdftex]{graphicx}
\usepackage{tikz}
\usepackage{algorithm}
\usetikzlibrary{positioning, fit, arrows.meta, shapes}
\usepackage{latexsym}
\usepackage[noend]{algpseudocode}
\usepackage{url}
\usepackage{soul}
\usepackage{amsmath}

\aclfinalcopy 

\usepackage{ifsym}

\title{Selecting Informative Contexts Improves Language Model Fine-tuning}

\author{Richard Antonello \\
UT Austin \\ 
\texttt{rjantonello@utexas.edu} \\\And
Nicole M. Beckage \\
Intel Labs \\
\texttt{nicole.beckage@intel.com} \\\AND
Javier S. Turek \\
Intel Labs \\ 
\texttt{javier.turek@intel.com} \\\And 
Alexander G. Huth \\
UT Austin \\
\texttt{huth@cs.utexas.edu} \\}

\begin{document}
\maketitle
\begin{abstract}
  Language model fine-tuning is essential for modern natural language processing, but is computationally expensive and time-consuming. Further, the effectiveness of fine-tuning is limited by the inclusion of training examples that negatively affect performance. Here we present a general fine-tuning method that we call \textit{information gain filtration} for improving the overall training efficiency and final performance of language model fine-tuning. We define the information gain of an example as the improvement on a validation metric after training on that example. A secondary learner is then trained to approximate this quantity. During fine-tuning, this learner selects informative examples and skips uninformative ones. We show that our method has consistent improvement across datasets, fine-tuning tasks, and language model architectures. For example, we achieve a median perplexity of 54.0 on a books dataset compared to 57.3 for standard fine-tuning. We present statistical evidence that offers insight into the improvements of our method over standard fine-tuning. The generality of our method leads us to propose a new paradigm for language model fine-tuning --- we encourage researchers to release pretrained secondary learners on common corpora to promote efficient and effective fine-tuning, thereby improving the performance and reducing the overall energy footprint of language model fine-tuning.
\end{abstract}

\section{Introduction}

Language modeling is the task of generating language from context. This is often framed as an autoregressive task, where a model predicts the conditional probability of the next word based on the sequence of previously observed or generated tokens. Language modeling has seen a recent surge in relevance thanks to its success as a pretraining objective for self-supervised representation learning. The most prominent language models today are Transformer-based models \citep{vaswani2017attention} such as BERT \citep{devlin2019bert} and GPT-2 \citep{radford2019language}. 

Language models are most commonly trained with backpropagation using traditional NLP loss functions such as cross entropy. These loss functions are designed so that the models are rewarded for assigning high probability to text that appears commonly in the training corpus. 
The energy and computational costs of training a state-of-the-art language model from scratch are very high, to the point of impracticality for most researchers. One recent estimate suggests that training a single state-of-the-art model with architecture search takes more energy than five cars will use in their entire lifetimes \citep{DBLP:conf/acl/StrubellGM19}. In practice, this cost is sidestepped by pretraining, where a language model is trained once and then released publicly. This language model can then be updated for use in other tasks through fine-tuning. For example, a generic language model can be fine-tuned to generate text that matches the style and syntax of any new corpus \cite{howard2018universal}. While better than training from scratch, the cost of fine-tuning such large networks is still relatively high. Fine-tuning to convergence for a single task can easily take in excess of a day on multiple energy-intensive GPUs \cite{DBLP:conf/acl/StrubellGM19}. 

Recent work analyzing the fine-tuning process has shown that it has high variability between runs and is particularly sensitive to data ordering \cite{dodge2020finetuning}. Those authors propose to overcome this variability by training models using many random seeds and then only keeping the best, effectively trading computational efficiency for model performance. While this improves performance, the reasons for the high variability between random seeds have yet to be explored. We hypothesize that much of this variability can be explained by the random selection of highly ``informative" training examples, which most effectively capture low-level distributional statistics of the target corpus.  If this is the case, then it should be possible to quickly screen for informative training examples, ensuring high performance at reduced cost.

In this paper, we suggest replacing the \textit{retrospective} approach of testing many random seeds \cite{dodge2020finetuning} with a \textit{prospective} approach to improving the effectiveness of language model fine-tuning. Our approach uses a secondary learner to estimate the usefulness of each training example, and then selects only informative examples for training. We show that this technique works well and is applicable in a variety of fine-tuning settings. We examine why it works well, and present evidence that supports our hypothesis about informative examples explaining data ordering effects. In addition to performance gains, this method may mitigate the energy impact of deep neural network language modeling, as we require fewer backpropagation steps than other techniques that trade computational power for fine-tuning performance. 

\section{Related Work}

Several methods have recently been proposed to improve language model fine-tuning performance. \citet{Lee2020Mixout:} proposed a technique based on neural network dropout \cite{dropout} for regularizing finetuned language models that involved stochastically mixing the parameters of multiple language models for the same domain, and further demonstrated the usefulness of pre-trained weight decay over conventional weight decay for improving language model fine-tuning performance. \citet{phang2018sentence} showed that adding supplementary training to pretrained language models using supervised tasks yielded state of the art results for BERT \cite{devlin2019bert}. \citet{moore-lewis-2010-intelligent} proposed a related technique for increasing the amount of language model training data from out-of-domain data sources that relies on filtering out high cross-entropy contexts as measured by an in-domain language model. \citet{tenney2019you} and \citet{liu2019linguistic} have both suggested that language model finetuning is better in Transformer-based models when starting when using features from intermediate layers as opposed to later layers.

The instability of language model fine-tuning has previously been investigated by others. \citet{mosbach2020stability} suggested that this instability is caused by a combination of insufficiently general training sets and optimization challenges. \citet{zhang2020revisiting} investigated how similar factors, such as non-standard optimization techniques and overreliance on a standard number of training iterations hurts the performance of fine-tuned language models.  \citet{dodge2020finetuning}, whose work we replicate and build on here, showed that language model finetuning is sufficiently stochastic so that even random seed searches are a suitable technique for improving their overall performance. 

\section{Background}
A language model $L$ is a function with parameters $\theta$, which, when given an ordered sequence of tokens $X = \{x_1,\dots,x_n\}$ as input, outputs a probability distribution over the next token $y$:
\begin{equation*}
L(X;\theta) = \hat p(y|X).
\end{equation*}
Given a test set $\mathcal{T}$ of (sequence, next token) pairs, $\mathcal{T}=\{(X_1,y_1),\dots,(X_n,y_n)\}$, the perplexity $\Lambda(\mathcal{T};\theta)$ of the language model $L(X;\theta)$ over the set $\mathcal{T}$ is defined as: 
$$\Lambda(\mathcal{T};\theta) = 2^{ -\sum_{(X_i,y_i) \in \mathcal{T}}  \bar p(y_i) \cdot \log_2  L(X_i;\theta)},$$
where $\bar p(y_i)$ denotes the one-hot probability distribution that assigns all of its probability mass to the token $y_i$. Autoregressive language models such as GPT-2 \cite{radford2019language} are trained to minimize perplexity using backpropagation on very large training corpora.

In practice, pre-trained language models are often fine-tuned using a new corpus or transferred to a new task \cite{howard2018universal}. Formally, let $\mathcal{F}=\{(X_i,y_i)\}_{i}$ be a target set. Fine-tuning on the set $\mathcal{F}$ tries to minimize the expected value of the loss function $\Lambda$:
\begin{equation}
    \hat{\theta} = \arg\min_{\theta}\mathbb{E}(\log_2\Lambda(\mathcal{F};\theta)).
    \label{eq:fine-tune-problem}
\end{equation}
The initial parameterization $\hat{\theta}_0$ of the language model is defined by its pre-trained  parameters $\hat{\theta}_0 = \theta$.
The fine-tuning problem in Eq.~\eqref{eq:fine-tune-problem} is then solved by applying stochastic gradient descent (SGD) on samples from $\mathcal{F}$. Namely, for a given batch $\mathcal{B}$ of samples from $\mathcal{F}$, the language model parameters are updated by $\hat{\theta}_k \leftarrow \hat{\theta}_{k-1} - \alpha \nabla\Lambda(\mathcal{B};\hat{\theta}_{k-1})$, where $\alpha$ is the step size. We refer to methods that randomly sample contexts to update pretrained model parameters as \textit{standard fine-tuning}.

While random sampling methods are useful \cite{bottou1991stochastic}, the stochasticity of context sampling suggests an avenue for additional improvement. Such methods make no assumption on the informativeness of the examples in $\mathcal{F}$, instead relying on randomness to find useful training samples. It is worth asking ourselves: can we efficiently measure the informativeness of an example? And if so, can we exploit that measurement for additional fine-tuning improvement?


\section{Information Gain Filtration}

\subsection{Informativeness of an Example}
Next, we characterize the informativeness of an example $(X,y) \in \mathcal{F}$, given a pre-trained language model $L(X;\theta)$ and a target dataset $\mathcal{F}$. We define an example $(X,y)$ as ``informative" if our estimate of the improvement that it will grant to the model exceeds a chosen threshold. Namely, if we expect that a given example will reduce model perplexity by more than a preset amount, then we will denote it as ``informative".

We define the \textit{information gain} (IG) of a example $(X,y)$ over an objective set $\mathcal{O}$ as the difference in perplexity measured on the objective set $\mathcal{O}$ before and after training on the example $(X,y)$,
\begin{equation}
IG_{\mathcal{O}}(X,y) = \Lambda(\mathcal{O};\theta'(X,y)) - \Lambda(\mathcal{O};\theta),
\label{eq:IG}
\end{equation}
where $\theta$ is the initial parameterization of the language model and $\theta'(X,y)$ is the parameterization after backpropagating the loss associated with training example $(X,y)$. 
The objective set $\mathcal{O} = \{(X_1,y_1),\dots,(X_n,y_n)\}$ is a held-out subset of training data that informs our decision about which contexts are informative. In practice, the objective set could be a subset of the fine-tuning set $\mathcal{F}$.
For brevity, we denote $IG_{\mathcal{O}}(X,y)$ as simply $IG(X)$ since there exists an implicit direct bijection between all $X$'s and $y$'s and the objective set is implied.


\subsection{Filtering Examples}
Since information gain evaluates the informativeness of an example, we next propose a method that exploits it for fine-tuning.
Let us assume that the method encounters a new example $(X,y)$. Then, the method has a choice between two actions:
\begin{itemize}
\item \textsc{Backprop}: update the language model parameters $\theta$ by backpropagating the loss $\Lambda(\{(X,y)\}; \theta)$, taking the gradient descent step, and updating parameters from $\theta$ to $\theta'$.
\item \textsc{Skip}: leave the language model parameters unchanged.
\end{itemize}
With this idea in mind we define the function\footnote{Due to its intuitive similarity with notions in reinforcement learning \citep{DBLP:journals/corr/MnihKSGAWR13} of using a network to approximate the expected value of a given action, we abbreviate this normalized informativeness metric as a ``$Q$-value" \cite{watkins1992q}} $q(X,action)$ and assign a value to each of the actions above:
\begin{eqnarray}
q(X, \textsc{Backprop}) &=& IG(X) \\
q(X, \textsc{Skip}) &=& T_{\textsc{Skip}}, 
\label{eq:Q_skip}
\end{eqnarray}
where $T_{\textsc{Skip}}$ is a free ``threshold'' parameter for deciding which $IG(X)$ values are sufficiently high to warrant backpropagation. 

Following this definition, we can apply a greedy policy for filtering examples during fine-tuning:
$$\pi(X) = \textbf{argmax}_{a \in \{\textsc{Backprop}, \textsc{Skip}\}} q(X, a).$$
By filtering examples in this way, we aim to reduce the effect of variability in data order observed in previous work \cite{dodge2020finetuning}, and improve the generalizability of our training set \cite{mosbach2020stability}. By doing this, we expect to improve the performance of our language model. 
We call this technique \textit{Information Gain Filtration} or simply \textit{IGF}. 



\subsection{Approximating Information Gain}
Thus far, we have described a general method to segregate informative from non-informative examples, deferring the issue of computational cost. Computing $IG(X)$ in Equation \eqref{eq:IG} entails a backpropagation step, making direct application of $q(X,action)$ at least as expensive as standard fine-tuning. 
To address this issue, we aim to approximate the information gain $IG(X)$ using a separate model that we will call the \textit{secondary learner} and denote with $\hat{Q}(X)$. 

To train this secondary learner, we first construct a training dataset $\mathcal{D}$ by measuring $IG(X)$ for a random subset of examples drawn from the fine-tuning set $\mathcal{F}$. The objective set $\mathcal{O}$ used to compute $IG(X)$ is selected as a different subset of $\mathcal{F}$. Each entry in $\mathcal{D}$ consists of a pair of the input text $X$ and its associated $IG(X)$ value, i.e., $\mathcal{D} = \{(X_1, IG(X_1)),\dots,(X_n, IG(X_n))\}$. We then train the secondary learner $\hat Q$ to approximate a normalized $IG(X)$ given $X$. We normalize $IG(X)$ so that $T_{\textsc{Skip}}$ can be interpreted as a standardized threshold on the selectivity of the filtration.
Finally, the resulting secondary learner $\hat Q$ is used to filter examples during fine-tuning.
Algorithm \ref{alg:igf} summarizes IGF with a secondary learner for language model fine-tuning.

\begin{algorithm}
    \caption{Information Gain Filtration \label{alg:igf}}
     \textbf{Input:} Fine-tuning ($\mathcal{F}$) and objective ($\mathcal{O}$) dataset of contexts, $(\mathcal{X}, \mathcal{O}) := \{(X_1,y_1),...,(X_n,y_n)\}$,
     parameterization of initial pretrained LM, $\theta$, and initial secondary learner model $\hat{Q}$.
     
\textbf{Parameters:} Size of learner dataset, $s$, and \newline
    threshold parameter, $T_{\textsc{Skip}}$

\textbf{Output:} $\theta'$, new parameterization for the LM
    \begin{algorithmic}[1]
    \State Initialize $\mathcal{D}, \mathcal{B} = \{\}$. 
    \For{$i=0 \dots s$}
    \State Sample context $(X_i, y_i)$ from $\mathcal{X}$.
    \State Append $(X_i, IG_{\mathcal{O}}(X_i, y_i))$ to $\mathcal{D}$. 
    \EndFor
    \State Normalize $IG_{\mathcal{O}}(X,y)$ values in $\mathcal{D}$ to $\mathcal{N}(0,1)$.
    \State Train secondary learner $\hat Q$, using dataset $\mathcal D$. 
    \For{$i=0 \dots$number of batches$-1$}
    \While{$|\mathcal{B}| <$ batch size}
     \State Sample context $C = (X, y)$ from $\mathcal{X}$.
     \If{$\hat Q(C) \geq T_{\textsc{Skip}}$}
     \State Add $C$ to batch $\mathcal{B}$.
     \EndIf
    \EndWhile 
    \State Backpropagate over batch $\mathcal{B}$, updating $\theta$
    \State Reset batch $\mathcal{B} = \{\}$.
    \EndFor
    \State Return $\theta$.
    
    \end{algorithmic}

\end{algorithm}

\subsection{Scheduled Thresholding}
The secondary learner training set $\mathcal{D}$ is constructed using the initial pretrained model parameters $\theta_0$. This means that the effectiveness of the learner at distinguishing ``high quality'' from ``low quality'' examples should degrade as the parameters diverge from their initial values. 
To ameliorate this problem, Equation \eqref{eq:Q_skip} can be modified by changing $T_{\textsc{Skip}}$ during the fine-tuning process. Since $\hat Q$ is most accurate at the first step, we scheduled $T_{\textsc{Skip}}$ to switch from highly selective (a high value) to highly permissive (a low value). This allows the model to take advantage of the accurate predictions for $IG(X)$ early in the fine-tuning process without overfitting once those predictions become less accurate later on.

\section{Results}

Here we first provide an empirical analysis suggesting that IGF outperforms standard fine-tuning across different choices of datasets, fine-tuning tasks, and neural architectures. We follow this analysis with an examination of why IGF works, and an exploration into the statistical properties of standard fine-tuning and of IGF. We tested these results on a standard Books dataset \cite{zhu2015aligning}, a ``mixed" dataset which is composed of training examples from two corpora (the Books corpus and a corpus of scraped Reddit comments \cite{huth2016natural}), and the WikiText-103 dataset \cite{merity2016pointer}. The Books corpus allows us to fairly compare standard fine-tuning against IGF, whereas the Mixed corpus allows us to analyze the effectiveness of the method at separating informative contexts from uninformative ones. 

In practice, our secondary learner, $\hat Q$, represents the input text $X$ by embedding it with 768-dimensional byte-pair embeddings \cite{gage1994new}. We then pass the input representations through a convolution with kernel width 3, followed by max-pooling operation over the time axis and a 2-layer feedforward network. This architecture was refined through coordinate descent, and evaluated on a separate held-out set of measured $IG(X)$ values. The choice of architecture does not strongly affect method performance (see Appendix A, Figure \ref{archinv}). Additionally, a neural network is not necessary for the learner, as simpler learning methods are sufficient (see Figure \ref{learners}). 

\begin{figure}[htb!]
\centering

\includegraphics[width=1.0\linewidth]{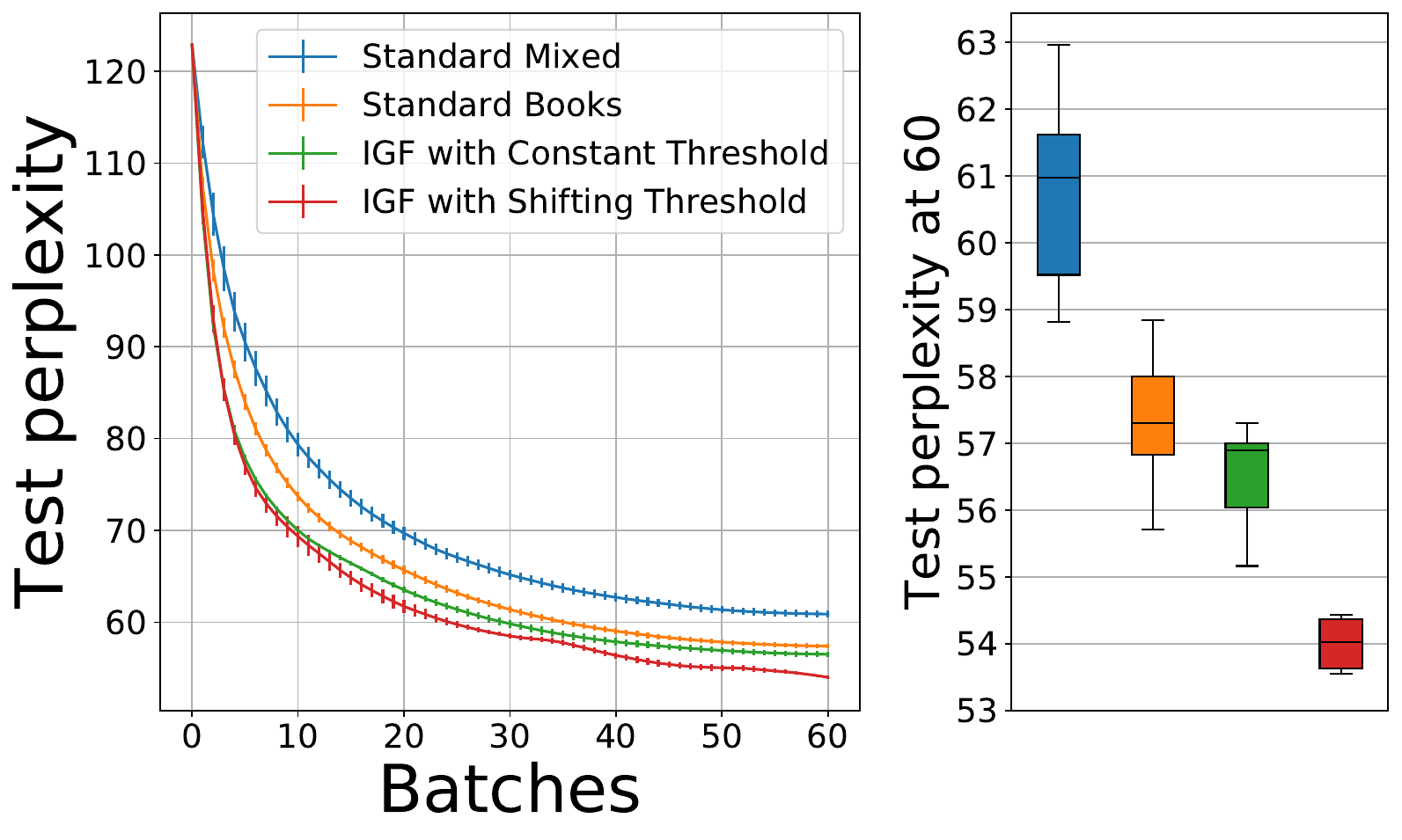}
\caption[var]{\textit{Comparing IGF to Standard Fine-tuning:} IGF with constant ($p < 10^{-3}$, $t$-test) and shifting ($p < 10^{-6}$, $t$-test) thresholding significantly outperform standard fine-tuning. The left-hand figure shows test-set perplexity after each fine-tuning batch, averaged over 50 runs (error bars denote $\pm$ one standard error). The right-hand figure shows the perplexity of each method after 60 batches. IGF with shifting thresholding (red) clearly improves over standard batched fine-tuning with Adam. For the constant threshold, $T_{\textsc{Skip}}$ was set to 0.75. For the shifting threshold, $T_{\textsc{Skip}}$ was change from 1 to -1 after the tenth batch.  In both IGF tests, the Mixed corpus was used and a set of 160 example contexts of 32 tokens each from the Books corpus was used as the objective set.}

\label{sameset}
\end{figure}

\subsection{Language Model Fine-tuning}

We first compare IGF directly to standard fine-tuning, which we define as basic batched stochastic gradient descent with Adam \cite{DBLP:journals/corr/KingmaB14} using random samples from the target corpus.  For initial tests, we chose the pretrained GPT-2 Small Transformer model, a commonly used unidirectional language model with roughly 124 million parameters. We used the publicly available GPT-2 Small implementation of the \texttt{transformers} package \cite{Wolf2019HuggingFacesTS}.  We performed 50 runs each of standard fine-tuning on (1) training examples sampled from the Mixed corpus, and (2) from the easier Books corpus. We then performed 50 runs of IGF using two thresholding schedules, one with a fixed $T_{\textsc{Skip}}$ and one with shifting $T_{\textsc{Skip}}$. For both methods, batches of size 16 were used to train the language model with a learning rate of $5 \times 10^{-5}$ and $\beta_1=0.9, \beta_2=0.999$. The convolutional network that we used for our secondary learner was trained using SGD with Adam with a learning rate of $10^{-5}$ and $\beta_1=0.9, \beta_2=0.999$. Both types of IGF runs were performed on the strictly more challenging Mixed corpus only. In all cases model perplexity was tested on a set drawn solely from the Books corpus. Figure \ref{sameset} plots the averaged fine-tuning curves of these 4 different approaches over 60 batches. We see that IGF significantly improves final test perplexity when compared to standard fine-tuning on both the Mixed corpus and the Books corpus. Standard fine-tuning on Books achieves a median perplexity of 57.3, compared to 56.9 for IGF with a constant threshold and 54.0 for IGF with the shifting threshold schedule.\footnote{Demo code and data can be found at \url{https://github.com/huggingface/transformers/tree/main/examples/research_projects}.} \emph{All 50 runs of IGF with a shifting schedule outperformed all 50 standard fine-tuning runs.} This means that the overall improvements to data order that IGF achieves through selective sampling of informative contexts are far in excess of what might be reasonably achieved through random sampling of contexts.

\begin{figure}[htb!]
\centering

\includegraphics[width=1.0\linewidth]{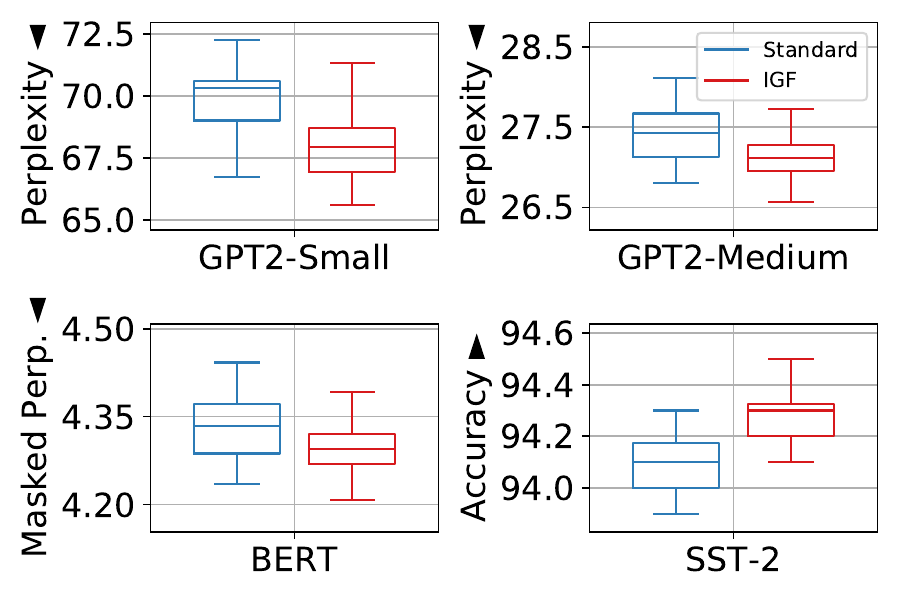}

\caption[var]{\textit{IGF is Invariant to Model Variation:} We compare performance of IGF and standard fine-tuning across a variety of choices of model specification and dataset. Box plots show results from 50 runs with each method. \textit{Top left:} IGF outperforms standard fine-tuning with an average test perplexity of \textbf{67.8} compared to \textbf{69.8} when fine-tuning on GPT2-Small. \textit{Top right:} When using the GPT2-Medium pretrained model, IGF converges to \textbf{27.1} as opposed to \textbf{27.4} for standard fine-tuning. \textit{Bottom left:} When fine-tuning BERT (a bi-directional language model trained to minimize masked perplexity rather than next-word perplexity), masked perplexity declines from \textbf{4.33} to \textbf{4.29}. \textit{Bottom right:} When fine-tuning instead to the Stanford Sentiment Treebank, a sentiment analysis task, IGF improves accuracy from an average of  \textbf{94.06} to \textbf{94.27}. The WikiText-103 dataset was use for all comparisons except for SST-2. All other model parameters are as in Figure \ref{sameset} and use a shifting thresholding schedule.  When fine-tuning on BERT and SST-2, the plotted metrics (masked perplexity and accuracy) were used instead of next-word perplexity to compute $IG(X)$. All differences are statistically significant to $p < 10^{-3}$.}
\label{model_variations}
\end{figure}

Next, we show that the improvements offered by IGF persist across several choices of dataset, fine-tuning specifications, and model architecture. Figure \ref{model_variations} shows the final converged values for fine-tuning GPT-2 Small on a different dataset from Figure \ref{sameset} (WikiText-103), a different architecture (GPT2-Medium), a different embedding space with different directionality (BERT) \cite{devlin2019bert}, and a different overall fine-tuning task (SST-2) \cite{socher2013recursive}. In every case, IGF  exceeds the performance of standard fine-tuning. This suggests that IGF is a resilient method that is broadly applicable to a variety of fine-tuning modalities and domains.

\begin{figure}[htb!]
\centering

\includegraphics[width=0.91\linewidth]{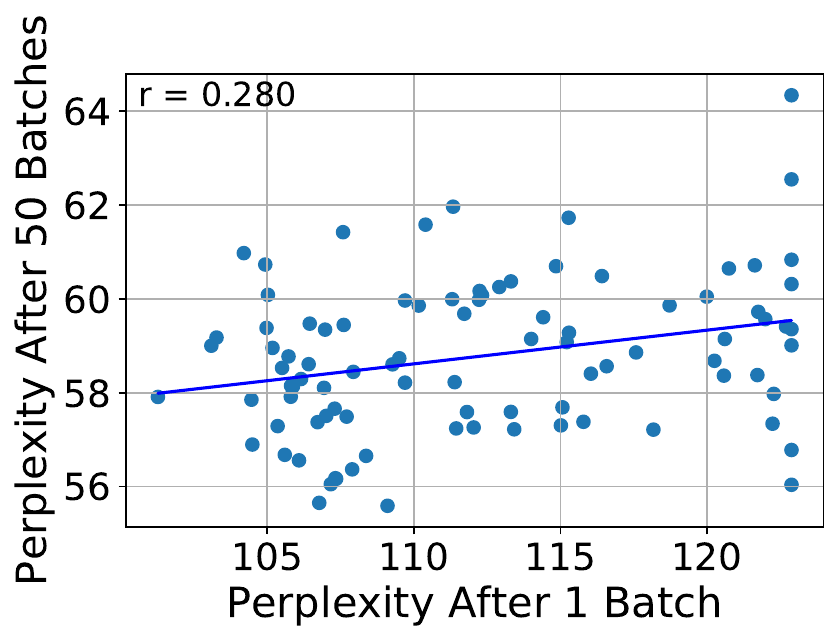}

\caption[naiveimpact]{\textit{Reduction in Perplexity in Early Steps is Predictive of Total Reduction}:  If the first batch in a fine-tuning run leads to a large reduction in perplexity, the fine-tuning run as a whole will tend to converge to a lower value ($r=0.28$). This is significant to $p<0.01$.}

\label{corr1}
\end{figure}

\subsection{Understanding IGF}


\begin{figure}[htb!]
\centering

\includegraphics[width=1.0\linewidth]{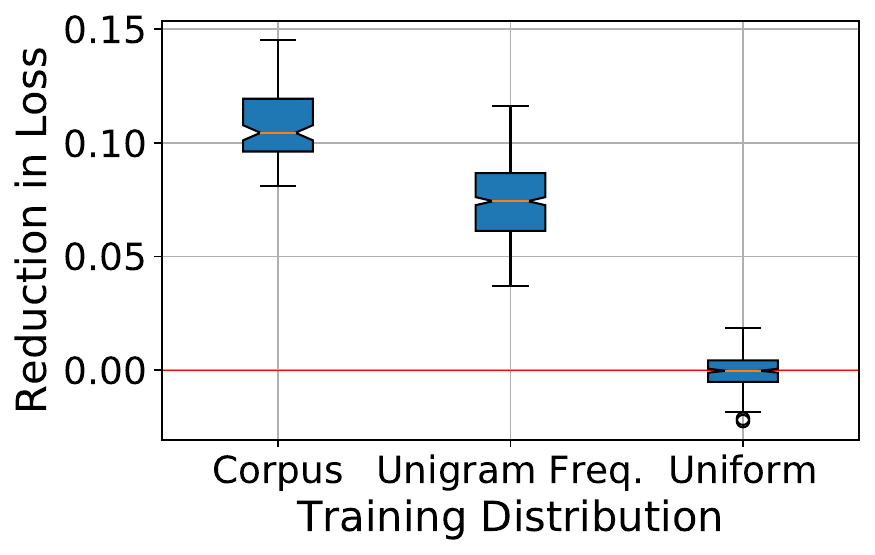}
\vspace{-2.0em}
\caption[dtfigcaption]{\textit{Learning the New Unigram Frequency Distribution Constitutes Most of the Benefit of Fine-tuning}: These plots show the reduction in cross-entropy of a GPT-2 language model, tested on a Reddit corpus after training on each 32 token contexts sampled from different distributions. Each example consisted of a word along with the preceding 32 words of context. Positive values indicate that learning from that example resulted in reduced loss on the test dataset. (left) \textit{Actual sequence from corpus.} The language model learns something useful from every example when finetuned on text from the corpus. (middle) \textit{Random sequence with preserved word probabilities.} For this sequence, 32 tokens are sampled to generate a context using the unigram probabilities for the Reddit corpus. Here the model also learns something useful from every example, despite being finetuned on scrambled text. (right) \textit{Random sequence with uniform word probabilities.} When the unigram probability distribution is replaced with a uniform probability distribution, the model no longer consistently learns. All pairs of distributions are different with $p<10^{-6}$.}
\vspace{-0.5em}
\label{dtfig}
\end{figure}

\begin{figure*}[htb!]
\centering

\includegraphics[height=3.5cm,width=0.7\linewidth]{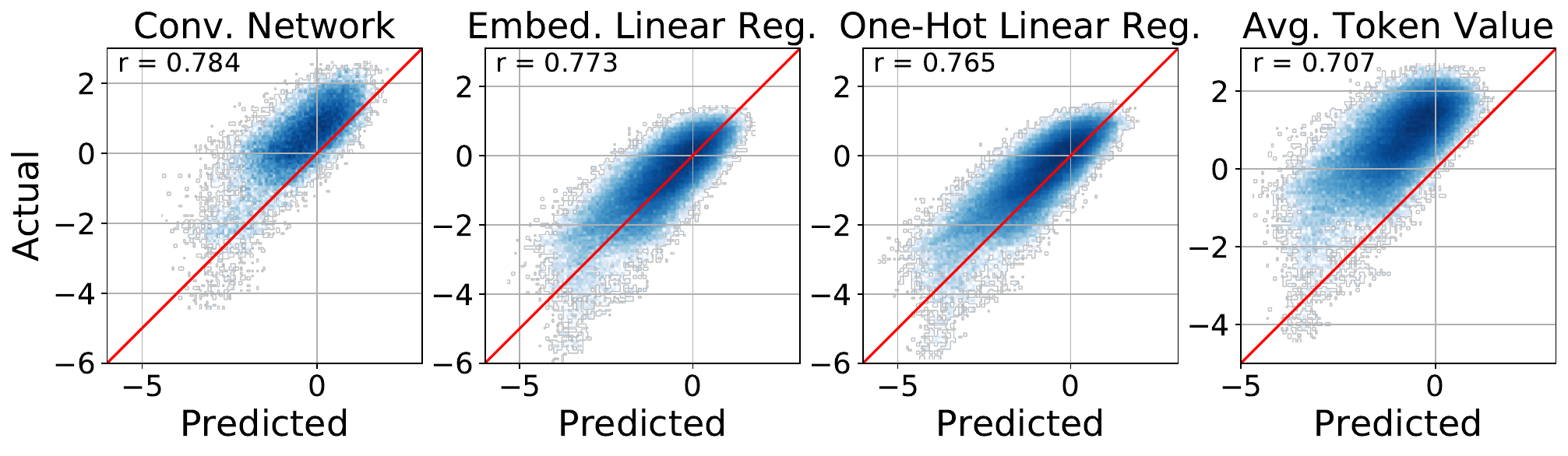}
\raisebox{1.0em}{\includegraphics[width=0.25\linewidth]{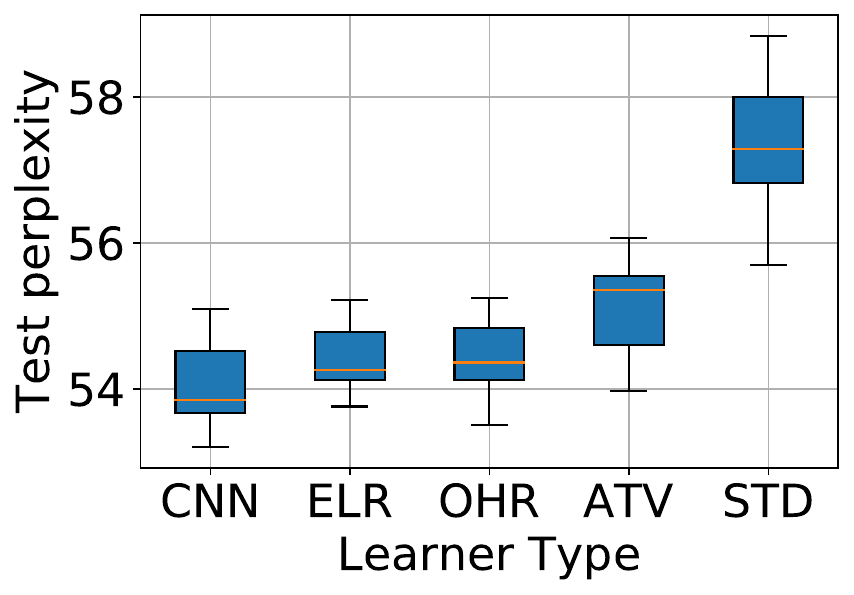}}
\caption[dtfigcaption]{\textit{Comparing The Ability of Simple Learners To Estimate Information Gain}: The above plots show the prediction accuracy (\textit{scatter plots on the left}) and overall fine-tuning performance of each learner when used during IGF (\textit{boxplot on the right}) for a variety of secondary learners. Each performs well in estimating $IG(X)$ when trained on a dataset of $(X, IG(X))$ pairs. The convolutional network  (\textit{far left}) which we chose as our secondary learner moderately outperforms the other simple learners.  As alternative learners, we also tested linear regression where $x$ is represented as its average embedded representation in the GPT-2 byte-pair embedding space (\textit{center left}), linear regression where $x$ is represented as a one-hot encoding over the token values (\textit{center}), and a trivial learner which estimated the value of a context as average of the values of the tokens that compose it, whose values are in turn computed as the average value of the training contexts they occur in (\textit{center right}). A comparison to standard finetuning without IGF (\textit{far right}) is included. As a difference of means, the CNN is statistically different ($p < 0.001$) from the other types of learners. For the one-hot and average token value learners, contexts with tokens appearing in the training set and not in the test set were excluded. All learners were trained on a dataset of 10,000 training examples.}
\label{learners}
\end{figure*}

It is clear that IGF is successful as a general method for improving fine-tuning performance, however \textit{why} this is the case remains unexamined. Here, we present an analysis of the statistical properties of fine-tuning that illuminates why IGF is able to improve over standard fine-tuning. 

A main assumption of IGF is that it is possible to approximate $IG(X)$. If $IG(X)$ is not approximable, then the secondary learner could not effectively filter out uninformative contexts and therefore would be useless. In order to support this assumption, we will first show that a given example is worth learning from even if it only possesses the correct low-level features of informative contexts, such as the correct unigram frequency distribution. We performed an experiment in which we fine-tuned a language model on either (1) real example sequences from a corpus, (2) artificial sequences that were constructed by independently sampling each token from the frequency distribution of the corpus, and (3) sequences constructed by uniformly sampling tokens from the set of all possible tokens. We then measured the change in loss on a separate portion of the corpus. Figure \ref{dtfig} shows the results of this experiment. The average reduction in loss for examples constructed using the unigram frequency distribution is significantly better than random and roughly 70\% as good as using real examples from the corpus. Thus, a significant fraction of the benefit of training on real contexts can be estimated by merely knowing the unigram frequency distribution from which those contexts were derived, which is easily estimable without knowing the particular parameterization of the language model itself. Therefore, it makes sense that IGF can inexpensively estimate whether a given context generalizes well to the target corpus.

The secondary learner only bases its estimates on the update to loss after the first backpropogation step. We might question whether early improvement translates to long-term improvement over the course of fine-tuning. If it did not, then the estimates that the secondary learner produces would eventually disappear as fine-tuning continued. \citet{dodge2020finetuning} observed that the quality of a fine-tuning run could usually be established by looking at the trajectory of the loss curve very early during training. In order to explain why these early estimates are sufficient for sample filtration, we attempted to determine whether training on good contexts early is an important element of the variability in data order between fine-tuning runs. Figure \ref{corr1} compares \textit{test perplexity} after training from a randomly sampled first batch against the test perplexity after many randomly sampled batches. Good early batches improve the probability of converging to an ideal final value. The correlation between the test perplexity after a single batch and the test perplexity after 50 batches, which is near convergence for most runs, is statistically significant ($r=0.28$). While this value appears somewhat low, it is significant and therefore can be exploited for improvements in performance.  
 
 \begin{figure}[htb!]
\centering
\includegraphics[width=1.0\linewidth]{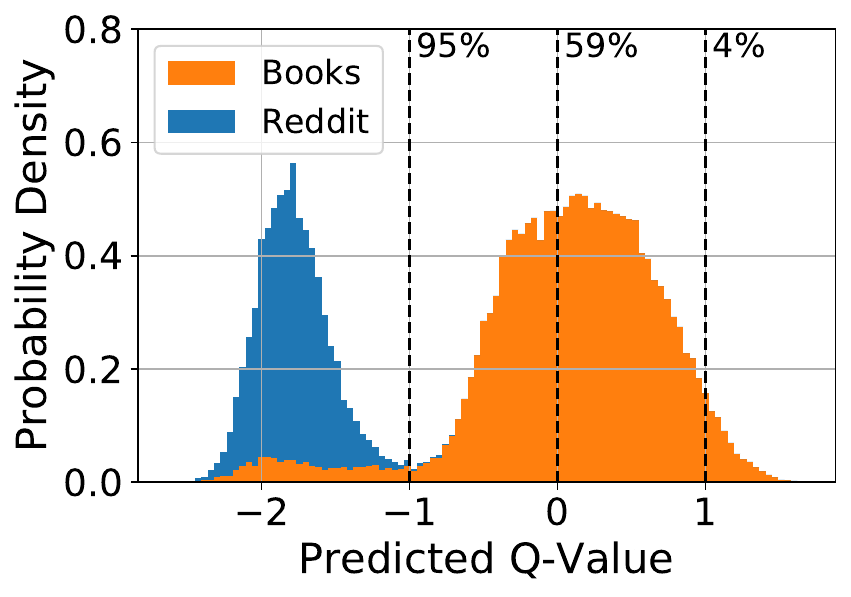}
\caption[corporaarediffcaption]{\textit{Normalized Predicted Q's by Training Corpus}:  In the mixed setting, a corpus composed of Reddit comments \textit{(25\% of contexts)} and a corpus of books \textit{(75\% of contexts)} were mixed into a single training dataset. Using the predicted $q$-value generated from our convolutional secondary learner, we can achieve good separation of the corpora using the information gain metric despite computing the true $q$-value using a small objective set. The percentage of examples from the Books corpus that are higher than several frequently referenced $T_{\textsc{Skip}}$ values are given for our dataset.}

\label{corporaarediff}
\end{figure}
 
Taken together, the pair of observations that \textit{(1)} early data quality is important, and \textit{(2)} that the quality of a context can be summarized by its low-level statistics serves to motivate our understanding of why IGF is effective. Specifically, if we can carefully ensure that early batches are good, as IGF does, then we will likely end up with a superior model after convergence.

\subsection{Understanding the Secondary Learner}

This raises the question of which contexts are considered ``informative"  by the secondary learner. To answer this question, we apply IGF to the Mixed corpus containing both Reddit and Books. We created a dataset of 10,000 $(X, IG(X))$ pairs using an objective set of 160 contexts with 32 tokens each drawn solely from the Books corpus. We used this dataset to train a secondary learner.  Next, the secondary learner was fed randomly sampled contexts from the Mixed corpus. Because the objective set contains only examples from one corpus, we expect the secondary learner to assign higher $IG(X)$ values to other examples from the same corpus. Figure \ref{corporaarediff} shows that there is indeed a significant difference in the distributions of $\hat Q$ values between the two corpora, demonstrating that the Books and Reddit corpora can be separated by the secondary learner. Almost all examples from the Reddit corpus are expected by the secondary learner to produce a reduction in perplexity that is at least one standard deviation below the mean. This indicates that the secondary learner can identify with strong confidence that Books corpus examples are more informative for fine-tuning towards the Books objective than Reddit corpus examples. It is also worthwhile to note that the secondary learner achieves dataset separation despite having access to just 160 labeled examples of 32 tokens in our objective set, a total of just 5120 tokens from the Books corpus, and zero examples from the Reddit corpus.

\setlength{\textfloatsep}{20pt}

\begin{figure}[htb!]
\centering

\includegraphics[width=1.0\linewidth]{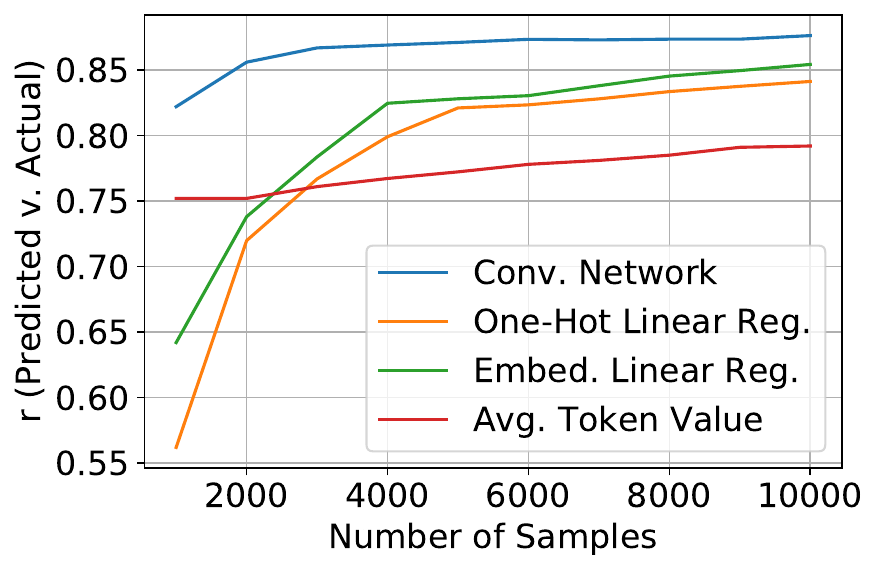}

\caption[var]{\textit{Comparison of the Sample Efficiency of Secondary Learners}: Here we compare the relative sample efficiency of the various secondary learners that were tested in the paper using contexts from the WikiText-103 dataset. We plot the correlation coefficient of the model prediction against ground truth as the number of samples in the training set for that model increases from 1000 to 10000. We see that the convolutional network, being the most highly regularized of the four models owing to its architectural structure and relatively low parameter size, is also the most sample efficient of all of the models tested.}
\label{Sampling}
\end{figure}

\subsection{Efficiency of IGF}

For previous results we used a simple convolutional neural network described in Section 3 as our secondary learner. However, it may not be necessary to use a such a complex model for $IG(X)$. Alternative methods could provide similar performance at less cost. Figure \ref{learners} shows predicted vs. actual normalized $IG(X)$ values for several learning methods. While the 45,000 parameter convolutional neural network is most effective at approximating $IG(X)$, other learners perform almost well. We encoded the contexts both by using the standard GPT-2 Small word embedding and with a one-hot encoding of the token identities. Standard linear regression performed on both encoding types (30K parameters for word embeddings and 450,000 parameters for one-hot encoding) performs nearly as well at approximating $IG(X)$ with a convolutional model. We also tested an even simpler learner with only 25,000 parameters that assigned each token a value by averaging the $IG(X)$ values for contexts that contained that token. Values for new contexts are then computed as the average of token values contained in that context. Even this model is a reasonable approximator of $IG(X)$. This underscores that, while $IG(X)$ is an extremely complex function to compute \textit{exactly}, it can nevertheless be effectively \textit{approximated} through simple unigram information. Figure \ref{Sampling} compares the performance of these secondary learners architectures across different numbers of training examples. Here the convolutional network is the most sample efficient method, as it can effectively learn $IG(X)$ with as few as 2,000 training examples. 

\subsection{Comparison to Random Seed Search}
\begin{figure}[htb!]
\centering

\includegraphics[width=1.0\linewidth]{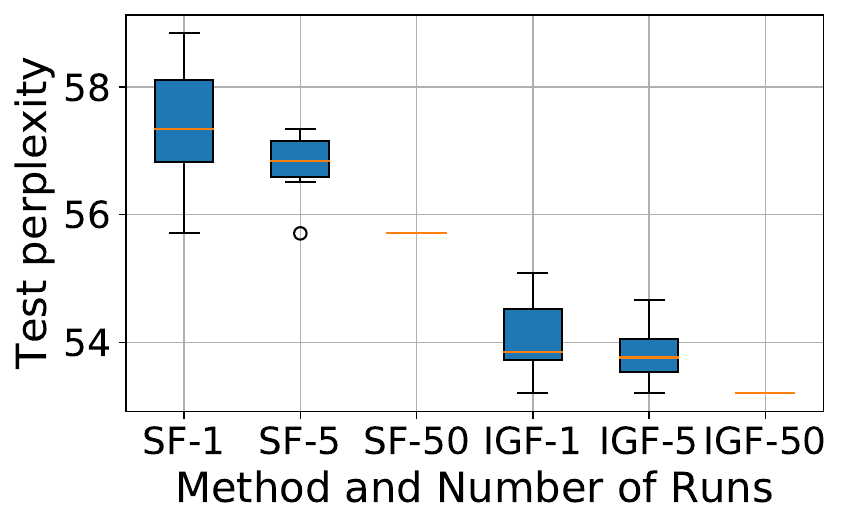}

\caption[var]{\textit{Prospective IGF Is More Efficient Than Retrospective Random Seed Search}: We show boxplots of the best run from differently-sized sets of runs to visualize the expected benefit of using random seed testing \cite{dodge2020finetuning} and compare it to the benefit of using IGF. Even one IGF run is significantly more effective than 50 random seed tests using standard fine-tuning, denoted here as SF. We further observe that the improvements to data order that come from IGF are somewhat disjoint from the improvements to data order than come with random seed testing, so both approaches can be applied simultaneously for further perplexity reduction. Sets of runs of each size were generated by sampling without replacement from a pool of independent 50 runs for each method. For the 50 run case, the minimum over the entire pool of runs for each method is plotted instead.} 
\label{conveff}
\end{figure}

We proposed IGF as a prospective alternative to the random seed search approach suggested by \citet{dodge2020finetuning}. Since IGF aims to replaces random search with a directed search, we expect IGF to be significantly more efficient. Of course the methods can also be combined: IGF can be run many times with different data orders, and then the best model selected. In Figure \ref{conveff} we compare the \citet{dodge2020finetuning} method, where the best model is selected across 1, 5, or 50 runs of standard finetuning, to a similar setup where the best IGF model is selected across 1, 5, or 50 runs. We find that even in a single run, IGF significantly outperforms choosing the best of 50 runs of standard finetuning. Still, IGF performance can be improved even further by choosing the best result across 5 or 50 runs. This suggests that while IGF exploits some of the benefits that could be gained from ideal data ordering, there are still improvements to be made over IGF for further improving data order during language model fine-tuning.


\section{Conclusion and Future Work}

In the context of language model fine-tuning, we have shown that a secondary learner can efficiently and effectively distinguish between informative and uninformative training examples. This secondary learner can be used to select useful training examples in a technique we call Information Gain Filtration, leading to better model performance than standard fine-tuning. 
We encourage researchers to release pretrained secondary learners for frequently used corpora, in order to enable more effective finetuning and save energy. This would cut down the largest computational cost of applying IGF while retaining the performance improvements across the field. We have included several examples of open-sourced secondary learners in the supplementary material to promote this paradigm.

This work also raises several questions. Since our focus was on developing a lightweight technique, the most complex secondary learner we tested was a small convolutional network. Data efficiency during training could potentially be further improved by using a more complex model. The question of how far one could reasonably take a function approximator network for estimating information gain remains unexplored. 

Finally, we do not fully understand why improving performance on early training batches results better performance at convergence. Is this exclusively a property of language models, or do other networks and tasks exhibit this phenomenon?  Answering this question could lead to better optimization methods across many different fields.


\section*{Acknowledgments}

We would like to thank all of the people whose contributions, opinions and suggestions helped with the development of this project, especially Kaj Bostrom, Greg Durrett, Shailee Jain, and Vy Vo. This project was funded by a generous gift from Intel, and by the Burroughs-Wellcome Career Award at the Scientific Interface.


\bibliography{acl2020}

\bibliographystyle{acl_natbib}

\appendix

\clearpage
\setlength{\textfloatsep}{20pt}

\section{Supplementary Material}
\subsection{Miscellaneous Figures}

\label{sec:supplemental}

\begin{figure}[htb!]
\centering

\includegraphics[width=1.0\linewidth]{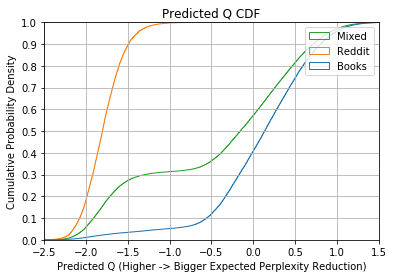}

\caption[var]{\textit{CDF of Predicted Q's}: CDFs of the datasets against the Books objective set. Note that a threshold of $T_{\textsc{Skip}} = -1$ almost entirely excludes contexts in the Mixed corpus that originated from the Reddit corpus. This allows IGF with a constant threshold of -1 on the Mixed dataset to perform almost identically to standard fine-tuning on just the Books corpus.}
\label{Tcdf}
\end{figure}

\begin{figure}[htb!]
\centering

\includegraphics[width=1.0\linewidth]{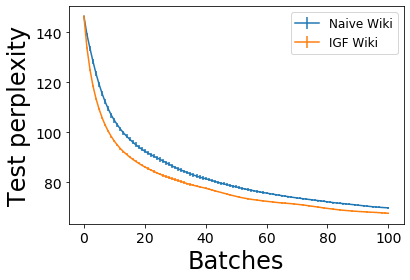}

\caption[var]{\textit{Replication of Performance Improvement on WikiText-103}: IGF significantly outperforms $(p < 10^{-4})$ standard fine-tuning without context filtering on the WikiText-103 dataset. We plot the model perplexity over many batches as in Figure 4 of the paper. This figure can be replicated by following the Jupyter tutorial provided along with the supplementary material.}%
\label{sameset_wiki}
\end{figure}

\begin{figure}[htb!]
\centering

\includegraphics[width=\linewidth]{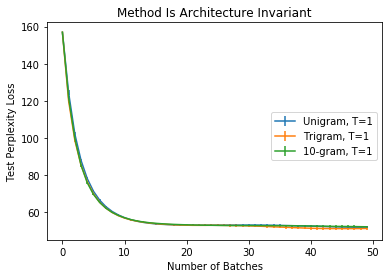}

\caption[archinvcaption]{\textit{Architecture Invariance}: The method performs similarly regardless of the convolutional setup of the model. Allowing the convolutional secondary learner to be informed by higher-order frequencies such as trigram and 10-gram do not significantly affect performance.}
\label{archinv}
\end{figure}

\begin{figure}[htb!]
\centering

\includegraphics[width=0.9\linewidth]{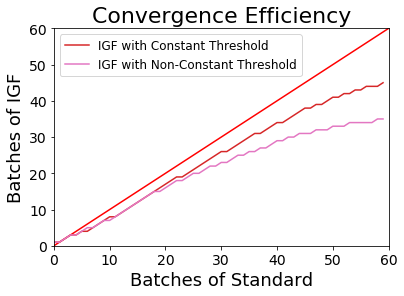}

\caption[var]{\textit{Improved fine-tuning Efficiency Over Standard fine-tuning: } We plot the number of batches it takes for each threshold schedule to exceed the perplexity of standard at each step.  This serves as a barometer for comparing the relative efficiency of fine-tuning. In the early stages of fine-tuning, we can see that IGF requires 30\%-40\% fewer backpropagation steps over standard fine-tuning. This suggests that IGF could be used as a more energy efficient alternative to standard language model fine-tuning.  Note that since IGF converges to a lower final value than standard fine-tuning, these values asymptote to a fixed value.}
\label{conveff2}
\end{figure}
\clearpage
\subsection{Sample High $IG(X)$ Contexts}
A few randomly sampled contexts from the Books corpora with $IG(X) > 1$ are given below. Note that all are highly structured conversations which are common of the narrative setting in the Books corpus: 
\begin{itemize}
    \item 't you. \newline He forced your hand with Max. " \newline " We're going to die, " she said. \newline " Aren't we?
    \item " The world is ending. " \newline " No it's not. " \newline Valerie snapped. \newline " It's a world war, that
    \item You? " \newline " Yep. \newline Did your dad leave? " \newline She nodded. \newline " They all said to tell you congratulations and they'll
\end{itemize}
\subsection{Sample Low $IG(X)$ Contexts}
A few sample contexts from the Books corpora with $IG(X) < -1$ are given below. Many of these contexts appear to be long, run-on sentences that are more challenging to follow:
\begin{itemize}
            \item n order ; you've got to make friends, you've got to put on a united front and for the governments of Earth that was no mean feat
            \item- headed eunuchs in crimson robes knelt in a cluster to one side of the dais, resting on their haunches and gazing at the woman and
            \item don't hold back, and by God, if I could be like you for even a moment, if I could have your strength, your courage, your
            \item frantically down one path, doubled back, and headed down another, like a frightened mouse trying to outsmart a determined cat in a warren of false trails and
\end{itemize}

\clearpage

\section{Iterated Information Gain Filtration}

Instead of scheduling the selectivity of the secondary learner to taper off as the fine-tuning process continues, we might instead \textit{replace} the learner periodically with a new learner trained on a new dataset of $(X, IG(X))$ pairs generated using the current parameterization of the language model. This process, which we call \textit{iterated infomation gain filtration} (IIGF), allows us to replace the obsolete learner that was trained to predict $IG(X)$ for early examples with a learner that is more relevant later in fine-tuning. IIGF has the added advantage of allowing us to keep $T_{\textsc{Skip}}$ high throughout fine-tuning, as secondary learner irrelevance is no longer a concern. This procedure is very computationally expensive, as the overhead in generating the new dataset and learner far exceeds the computational cost of fine-tuning.  Nonetheless, this enables finer control of data order throughout the fine-tuning process and further improvements in final perplexity over IGF with scheduled thresholding. Due to its computational expense, we ran a small set of 5 tests of iterated information gain filtration by training a secondary learner using a dataset built from example $(X, IG(X))$ pairs derived from a language model that had already been fully finetuned to the Books corpus. IIGF was able to improve these already-converged models by an average of 0.29 additional perplexity points after reconverging, with a standard deviation of 0.11 points.
\algrenewcommand\algorithmicindent{1.0em}%

 \begin{figure}
    
    \begin{algorithm}[H]
    \caption{Iterated Information Gain Filtration \label{alg:iigf}}
     \textbf{Input:} Training ($\mathcal{X}$) and objective ($\mathcal{O}$) dataset of contexts, $(\mathcal{X}, \mathcal{O}) = \{(X_1,y_1),...,(X_n,y_n)\}$, and
     
     parameterization of initial pretrained LM, $\theta$
     
\textbf{Parameters:} Size of learner dataset, $s$, \newline
    threshold parameter, $T_{\textsc{Skip}}$, and \newline
    number of batches per secondary learner reset, $t$

\textbf{Output:} $\theta'$, new parameterization for the LM
    \begin{algorithmic}[1]
    \State Initialize $\mathcal{D}, \mathcal{B} = \{\}$. 
    \ForAll{$i, 0 \leq i \leq num\_batches$}
    \While{$|\mathcal{B}| < batch\_size$}
    \If{$i \bmod t = 0$}
    \ForAll{$i, 0 \leq i \leq s$}
    \State Sample context $(X_i, y_i)$ from $\mathcal{X}$.
    \State Append $(X_i, IG_{\mathcal{O}}(X_i, y_i))$ to $\mathcal{D}$. 
    
    \EndFor
    
    \State Normalize $IG_{\mathcal{O}}(X,y)$ values in $\mathcal{D}$ to $\mathcal{N}(0,1)$.
    \State Train learner $\hat Q$, using $\mathcal D$ as a train set. 
     \State Sample context $C = (X, y)$ from $\mathcal{X}$.
     \If{$\hat Q(C) \geq T_{\textsc{Skip}}$}
     \State Append $C$ to batch $\mathcal{B}$.
     \EndIf
    \EndIf
    \EndWhile
    \State
    Update $\theta$ by backpropagating over batch $\mathcal{B}$, and clear batch $\mathcal{B}$.
    \EndFor
    \State Return $\theta$.
    
    \end{algorithmic}
    \end{algorithm}
    \end{figure}

\clearpage

\section{Relative Informativeness of Contexts}

\begin{figure}[htb!]
\centering
\includegraphics[width=1.0\linewidth]{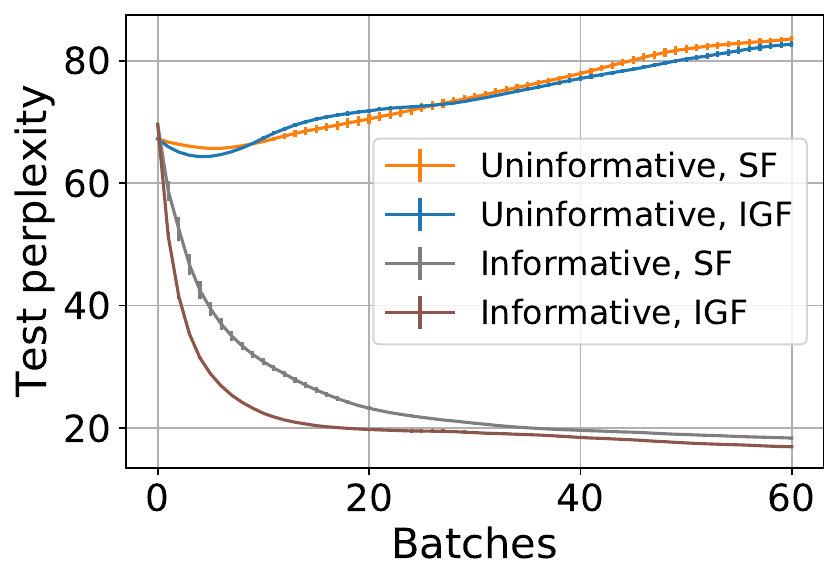}
\caption[corporaarediffcaption]{\textit{Informative Contexts Are Informed Contexts:} Shown above are plots of the evaluation performance of sets of 100 contexts rated as highly informative ($IG(X) > 1$) and uninformative ($IG(X) < -1$) by the secondary learner, as the language model is trained by either IGF or standard fine-tuning (SF). The contexts that the secondary learner rates as highly informative are also those contexts that the language model learns to predict very accurately after fine-tuning is complete. Conversely, contexts that the learner rates as poorly informative perform worse after fine-tuning. Examples of highly informative and poorly informative contexts from the Books corpus are presented in the supplementary material and support the assertion that the best contexts for fine-tuning are those that are highly predictable.}
\label{rel_info}
\end{figure}

Since our method uses a black box learner to estimate the informativeness of a given context, one might wonder what it is about these contexts that makes them more or less informative. To investigate this, we constructed test sets of 100 contexts each from the Books corpus which were rated as either highly informative ($IG(X) > 1$) or uninformative ($IG(X) < 1$) by the secondary learner. We then finetuned GPT2-Small using both standard fine-tuning and IGF as in Figure \ref{sameset} and periodically evaluated the performance of the model on the informative and uninformative contexts as training proceeded. Figure \ref{rel_info} shows that contexts which were rated as highly informative experienced a significantly greater reduction in perplexity over time as compared to contexts that were rated as uninformative. The poorly informative contexts actually performed worse on average after fine-tuning than either standard fine-tuning or IGF. This suggests that highly informative contexts are also \textit{highly informed}, or more easily predicted after fine-tuning on the target corpus.
Inspection of highly informative contexts shows that they tend to employ simple diction and basic sentence structure that is representative of the corpus, whereas uninformative contexts tend to employ complex sentence structure and atypical vocabulary. All highly rated contexts from the Books corpus consisted of dialog, which suggests that the secondary learner prioritizes linguistic patterns that are common to the fine-tuning corpus but rare in general writing. Since the Books corpus is composed of narrative stories heavy on dialog, it makes sense that conversations, which rarely appear in non-narrative corpora, would be rated as highly informative. The supplementary material gives some examples of highly informative and uninformative contexts from the Books corpus. 

\end{document}